\documentclass[letterpaper, 10 pt, conference]{ieeeconf}  

\IEEEoverridecommandlockouts                              

\usepackage{color}

\newcommand{\eg}{e.g.\ }
\newcommand{\Reffig}[1]{Figure~\ref{#1}}
\newcommand{\Refsec}[1]{Section~\ref{#1}}

\newcommand{\Reftab}[1]{Table~\ref{#1}}

\usepackage{graphics} 
\usepackage{epsfig} 

\usepackage{algorithm}
\usepackage{algpseudocode}
\usepackage{amsmath}
\usepackage{amssymb}
\usepackage{subfigure}
\usepackage{graphicx}
\usepackage{subfigure}
\usepackage{booktabs}
\usepackage{threeparttable}
\usepackage{multirow}
\usepackage{multicol}
\usepackage{makecell}

\title{\LARGE \bf
DSC: Deep Scan Context Descriptor for Large-Scale Place Recognition
}

\author{Jiafeng Cui$^{1, 4}$, Tengfei Huang$^{2, 3}$, Yingfeng Cai$^{2, 3}$, Junqiao Zhao$^{* 2, 3, 4}$, Lu Xiong$^{1, 4}$ and Zhuoping Yu$^{1, 4}$
\thanks{*This work is supported by the National Key Research and Development Program of China (No. 2018YFB0105103, No. 2018YFB0505400), the National Natural Science Foundation of China (No. U1764261, No. 41801335, No. 41871370)}
\thanks{$^{1}$School of Automobile, Tongji University, Shanghai, China}
\thanks{$^{2}$Department of Computer Science and Technology, School of Electronics and Information Engineering, Tongji University, Shanghai, China}
\thanks{$^{3}$The Key Laboratory of Embedded System and Service Computing, Ministry of Education, Tongji University, Shanghai, China}
\thanks{$^{4}$Institute of Intelligent Vehicles, Tongji University, Shanghai, China}
\thanks{*Corresponding Author:
        {\tt\small zhaojunqiao@tongji.edu.cn}}%
}

\begin{document}

\maketitle
\thispagestyle{empty}
\pagestyle{empty}

\begin{abstract}


LiDAR-based place recognition is an essential and challenging task both in loop closure detection and global relocalization. 
We propose Deep Scan Context (DSC), a general and discriminative global descriptor that captures the relationship among segments of a point cloud. 
Unlike previous methods that utilize either semantics or a sequence of adjacent point clouds for better place recognition, we only use raw point clouds to get competitive results. 
Concretely, we first segment the point cloud egocentrically to acquire centroids and eigenvalues of the segments. 
Then, we introduce a graph neural network to aggregate these features into an embedding representation. 
Extensive experiments conducted on the KITTI dataset show that DSC is robust to scene variants and outperforms existing methods. 

\end{abstract}

\section{INTRODUCTION}


Large-scale place recognition is an essential part of simultaneous localization and mapping (SLAM), which is widely applied in many autonomous robotics systems \cite{durrant2006simultaneous,stachniss2016simultaneous}. 
Visual place recognition has been extensively studied \cite{cummins2008fab,milford2012seqslam,salti2014shot,park2019robust}. 
However, these approaches are sensitive to illumination change and easily fail when the viewpoint of the input images differs from each other.
LiDAR-based methods are more robust to illumination change and viewpoint variants since the LiDAR sensor is capable of providing geometric structural information in a 360-degree view. 

\begin{figure}[htbp] 
        \centering 
        \large
        \includegraphics[width = 0.4\textwidth]{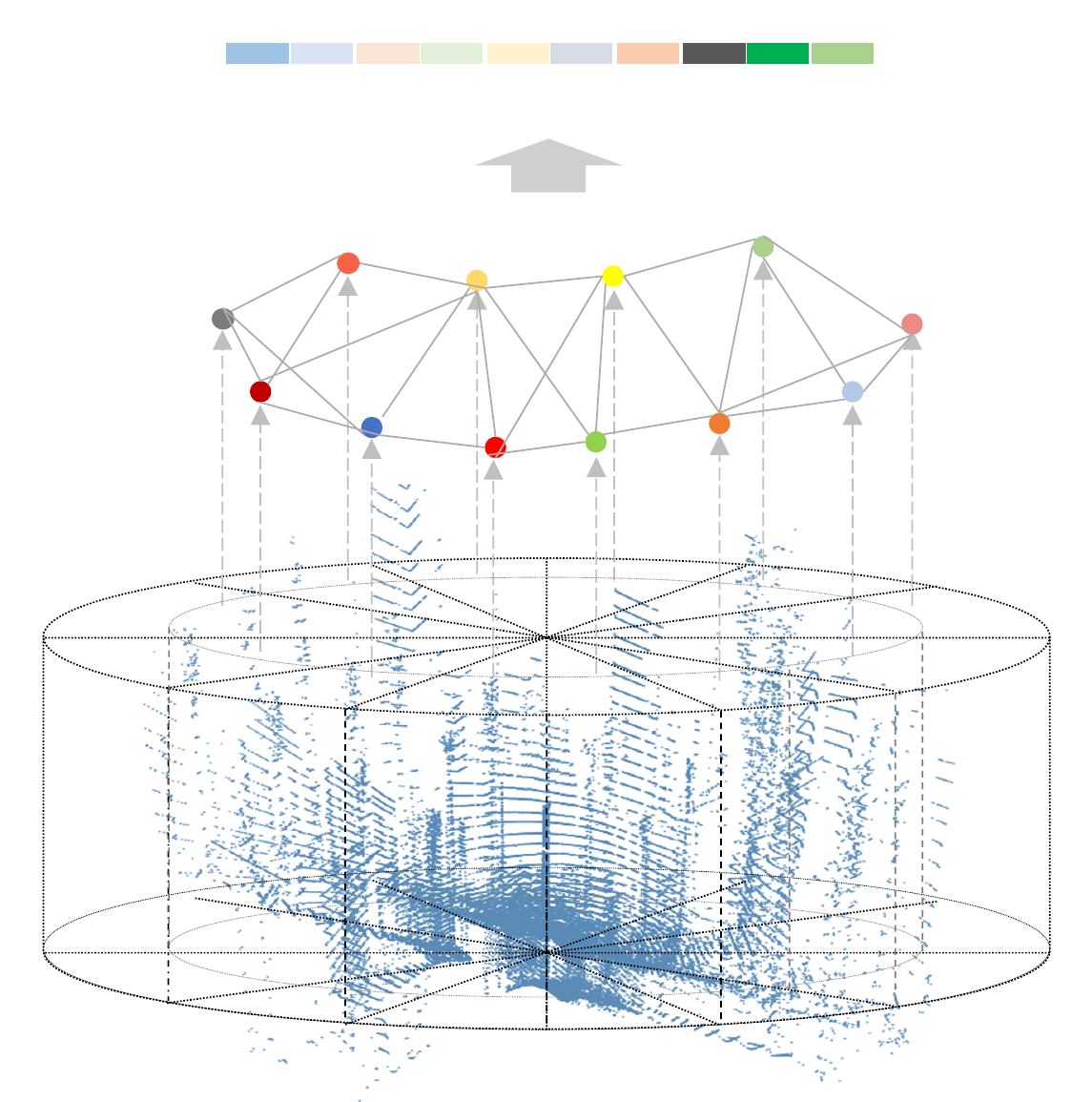} 
        \caption{
        A brief concept of Deep Scan Context. We first divide the point cloud into segments egocentrically, as a result, the segments will fall in certain segments. Then we extract the centroids 
        and eigenvalues of the segments. The centroids, as well as eigenvalues, will together be encoded into a graph neural network to finally output a global descriptor. }
        \label{overview}
\end{figure}

Previous works extracted local or global features from 3D point clouds \cite{bosse2013place,rusu2009fast,magnusson2009appearance,rusu2010fast,he2016m2dp,kim2018scan,uy2018pointnetvlad}. 
These pointwise methods pay little attention to potential topological information in the scenes, \eg{the arrangement of certain segments and their spatial relations}, hence it is difficult to distinguish scenarios with similar features but different structures.
In order to get distinct features, high-level representations such as geometrical segments or semantics were proposed \cite{dube2017segmatch,vidanapathirana2020locus,zhu2020gosmatch,kong2020semantic,li2021ssc}.
SegMatch \cite{dube2017segmatch} first used Euclidean clustering to segment a point cloud and search candidates by matching K-nearest neighbour (KNN) segments. 
Locus \cite{vidanapathirana2020locus} adopted the similar segmentation-based method and generate a global descriptor for a point cloud.
However, these Euclidean clustering-based methods performed poorly in a sparse point cloud.

SGPR \cite{kong2020semantic} introduced the semantic information to achieve a stable representation. 
SSC \cite{li2021ssc} further employed iterative closet point (ICP) and semantics to align the point cloud pairs to get more accurate matching results.
Although the semantic information is helpful to the task, it is difficult and time-consuming to get accurate semantic labels for generic scenes.

In this paper, we present a novel method, named Deep Scan Context (DSC), to effectively recognize similar places. 
As shown by \Reffig{overview}, given a point cloud frame, we first segment the raw data egocentrically. 
Compared with Euclidean clustering, it is more efficient and can avoid failure in sparse scenes \cite{vidanapathirana2020locus}. 
Next, we extract the centroid points of the segments and keep the topological relation between centroids using a graph neural network (GNN). 
To furture capture geometric features of the segments, we calculate eigenvalues of the segments to improve the saliency of description. 
These eigenvalues are then fed into another GNN to learn the relationships among features. 
Finally, the graph outputs are flattened into a fixed descriptor for computing candidates' similarities. 
Our main contributions in this paper are summarised as follows:

\begin{itemize}

\item We introduce an egocentric segmentation method that can better adapt to various scenes while other segmentation methods may fail to cluster enough segments.
\item We use the eigenvalues as features of the segments to achieve rotational invariance.
\item We propose a novel global descriptor to encode the structural features of these segments and their mutual topological features.

\end{itemize}

\begin{figure*} 
        \centering 
        \LARGE
        \includegraphics[width = 0.8\textwidth]{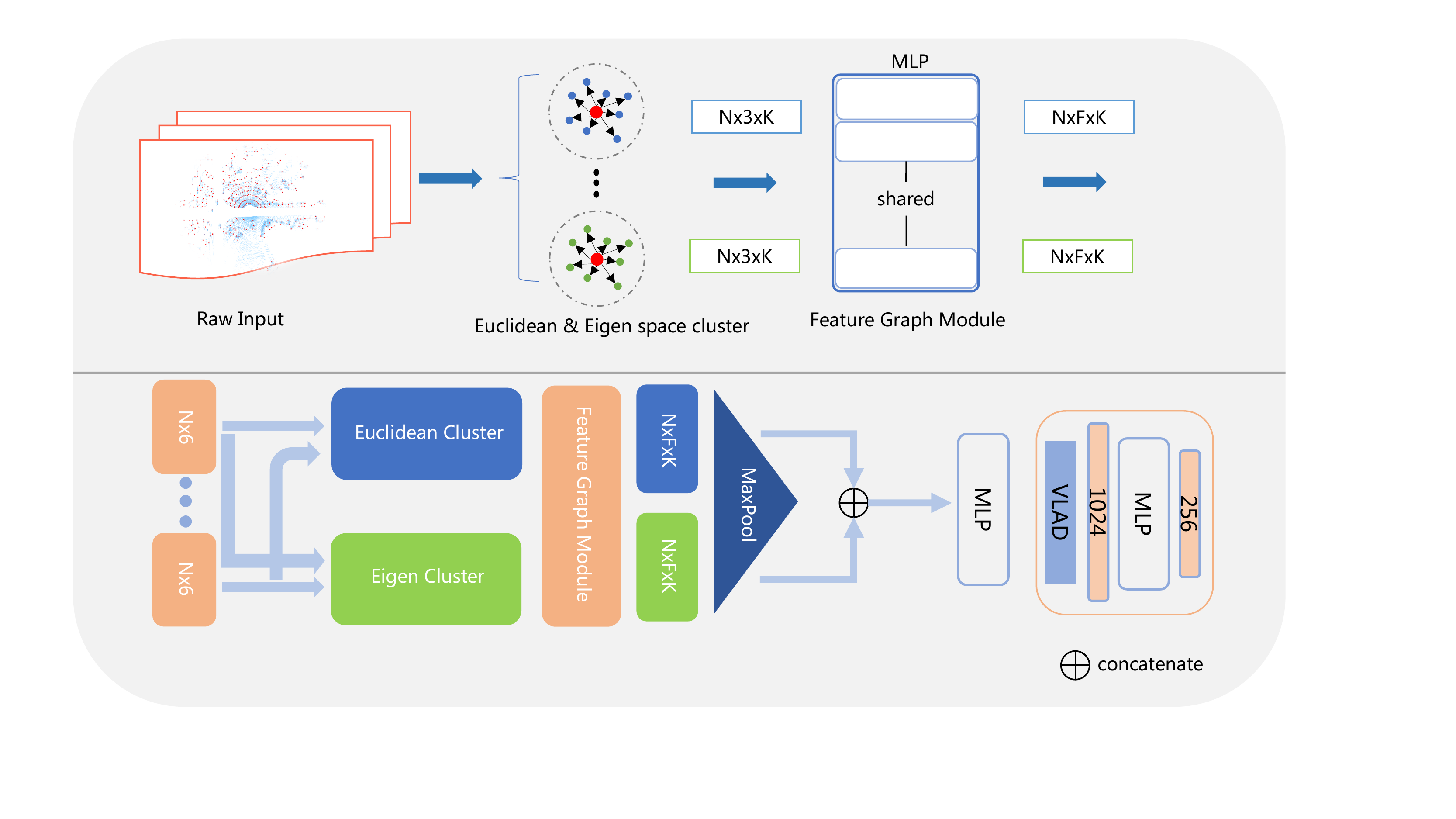} 
        \caption{
        The architecture of our method. Top: the illustration of each module in the method. Bottom: the illustration of the network that extracts Euclidean and Eigenvalue space features in parallel and concatenates them to aggregated features. 
        }
        \label{structure}
\end{figure*}

\section{RELATED WORK}

We give an overview of LiDAR-based place recognition methods including projection-based, handcraft descriptors-based, learning-based, and semantic-based methods.

\subsection{Projection-based methods}
Inspired by the success of visual place recognition methods \cite{mur2015orb,cummins2008fab,galvez2012bags}, many researchers proposed projection-based methods to utilize visual-based pipelines. 
M2DP \cite{he2016m2dp} projected the point clouds into different layers and collected the signature in each layer to generate a global descriptor. 
LiDAR-Iris \cite{wang2020lidar} extracted binary signature images from the raw point cloud and compared the Hamming distances with other candidates. 
Shan \cite{shan2021robust} projected the intensity of points into pixels to construct a grayscale image, converting it into a visual place recognition task. 
Disco \cite{xu2021disco} used a CNN network to extract features from scan context images \cite{kim2018scan} and got an initial orientation guess for later pose estimation. 
These projection-based methods lose the rich 3D information because of projection. 

\subsection{Handcraft descriptors-based methods}
Other works focus on extracting key points and combine the key points with their neighboring points to construct local features \cite{rusu2009fast}. 
Then the retrieval is implemented by vote-based or histogram-based approachs \cite{bosse2013place,steder2010narf}. 
Due to the poor repeatability of key points between frames, these methods are difficult to achieve highly accurate results.
SegMatch \cite{dube2017segmatch} first introduced a segmentation-based place recognition method with high-level features. 
They used the Euclidean clustering to segment the point cloud into a few segments and then compared each segment with its KNN neighbors in the presented map to extract corresponding candidates. 
Scan context \cite{kim2018scan} proposed a global descriptor that encoded the point cloud's max height values. 
This method is based on the hypothesis of structured scenes and may fail if there is no significant variation in height. 
ISC \cite{wang2020intensity} replaced height with remission information to describe the scene and adopted a fast coarse-to-fine match strategy. 
Similar to height, remission values are often affected by noise e.g. strong light and specular reflection. 

\subsection{Learning-based methods}
Based on PointNet \cite{qi2017pointnet}, PointNetVLAD \cite{uy2018pointnetvlad} extracted features for all the input points, then introduced the NetVLAD module to generate a global descriptor. 
LPD-Net \cite{liu2019lpd} further concatenated local features of each point to enlarge the receptive field and the retrieval performance improves a great margin. 
These pointwise methods take the whole point cloud into consideration,  and are sensitive to rotation variants. 
Locus \cite{vidanapathirana2020locus} segmented the point cloud through Euclidean clustering and learned the descriptor of each segment via a CNN model. 
To ensure the repeatability of the segmentation, Locus tried to find correspondences of the segments both in the spatial and temporal dimensions. 
The segments that had correspondences were kept and then integrated into a global descriptor to match other frames instead of matching segments independently. 
However, Euclidean clustering often fail to generate enough segments in sparse scenes where there are few features. 

\subsection{Semantic-based methods}
Chen \cite{chen2021overlapnet} designed a siamese network to acquire the overlap between pairs of LiDAR scans. 
The similarity of two point clouds can be measured by the overlap ratio.
Semantic information was first exploited in place recognition tasks to improve the ability of scene representation. 
SGPR \cite{kong2020semantic} and SSC \cite{li2021ssc} both used semantic information of the points to describe the scene.
SGPR segmented the point cloud in different semantic classes and used the Euclidean clustering in each class to get more instances. 
These semantic instances were represented as some nodes and the point cloud was represented as a graph.
SSC further introduced Iterative Closest Point (ICP) to align the two point clouds via semantic labels before computing the similarity score, which achieves the state-of-the-art performance. 
However, the predefined semantic labels are not easy to obtain, which limits the application of these semantic-based methods. 

To overcome the drawbacks above, we aim to design a generic and robust method that can be utilized in various scenes.

\section{METHODOLOGY}

In this section, we will introduce our method DSC in detail.
Our method consists of three main modules: egocentrically segmentation, segments feature extraction and deep scan context network. 

\subsection{Egocentrically Segmentation} 
Considering the nature of the rotating Lidar sensor, we first divide the 3D scan space into azimuthal and radial segments similar to \cite{kim2018scan}, which results in $N_s$ sectors along the azimuthal direction and $N_r$ rings along the radial direction.
\Reffig{overview} shows the egocentrically segmentation. 
Each segment is assigned an unique index number $\left(i,j\right)$, where $i \in \left \{ 1,\dots,N_s \right \}$ and $j \in \left \{ 1,\dots,N_r \right \}$. 
$N_s$ and $N_r$ are defined by the intervals of two dimensions. 
We set the radial interval as $\frac{L_{max}}{N_r}$ and the azimuthal interval as $\frac{2\pi}{N_s}$, where $L_{max}$ is the maximum detection range of the LiDAR. 
Specifically, given a point cloud $P \in R^{N\times3}$, for each point $p_m = \left(x_m,y_m,z_m\right)$, we calculate its polar coordinate $\left(\lambda{_m},\theta{_m}\right)$: 
\begin{equation}
        \small
\begin{aligned}
        \lambda _m = \sqrt{x_m^2 + y_m^2 } \\ 
        \theta _m = \arctan \left ( \frac{y_m}{x_m}  \right )
\end{aligned}
      \end{equation}
Then $p_m$ will be classified into segment $\left(i,j\right)$, if:
\begin{equation}
        \small
\begin{aligned}
        \left(j-1\right)\frac{L_{max}}{N_r} <= \lambda _m <= j\frac{L_{max}}{N_r} \\ 
        \left(i-1\right)\frac{2\pi}{N_s} <= \theta _m <= i\frac{2\pi}{N_s}
\end{aligned}
      \end{equation}
The point cloud is eventually divided into a set of segments $S = \left\{ S_1,\dots,S_t\right\}$. 
To get the topological relation of the segments, we calculate their centroids coordinates ($C = \left\{C_1,\dots,C_t\right\}$) and later feed them into a GNN as described in \Refsec{INTRODUCTION}.


\subsection{Segments Feature Extraction} 
Since centroid coordinates alone are not representive, more descriptions need to be extracted for the segments. 
SGPR \cite{kong2020semantic} used the pre-trained semantic segmentation network to classify the segments.
However, they are unable to distinguish the difference between segments of the same class. 
Segmap \cite{dube2018segmap} and Locus \cite{vidanapathirana2020locus} use a pre-trained CNN model on KITTI dataset to generate a descriptor for each segment. 
However, the results of segmentation on different datasets can be different, which leads to poor applicability of the pre-trained CNN model. 
Therefore, we calculate eigenvalues of the segments as the feature description which are not only rotation invariant \cite{xu2020geometry} but can also represent the shape information. 
Specifically, we first calculate the covariance matrices $V = \left\{V_1,\dots,V_t\right\}$ of all the segments. 
These matrices are symmetric positive definite as long as the segments are not flat \cite{xu2020geometry}.
The matrix can be decomposed by $V = RAR^T$, where $R$ is a rotation matrix and $A$ is the diagonal matrix. 
We denote the values on the main diagonal of $A$ as $\vec{\lambda}=\left(\lambda_1,\lambda_2,\lambda_3\right) \in \mathbb{R}$ which are the eigenvalues. 
For each point cloud, we can get a eigenvalues set $\Lambda = \left\{\vec{\lambda}_1,\dots,\vec{\lambda}_t\right\}$. 
More 2D and 3D shape features are proposed in \cite{weinmann2014semantic} that are derived from eigenvalues. 
However, they only focus on getting more representations on a single segment while we also consider relations of the features in the eigenvalue space.
Consequently, we choose the coordinates of centroids $C$ and the eigenvalues $\Lambda$ as the inputs of the deep scan context network. 

\subsection{Deep Scan Context Network} 
\label{sec_dscn}
As shown in \Reffig{structure}, our network mainly includes three parts: 1) Euclidean and eigenvalue space clustering, 2) feature graph construction, and 3) Net Vector of Locally Aggregated Descriptors (NetVLAD) \cite{arandjelovic2016netvlad}. 

\subsubsection{Euclidean and eigenvalue space clustering} 
For each input centroid, KNN search is utilized to cluster neighbor centroids both in the Euclidean space and the eigenvalue space.
Let $\left\{C_{i_1},\dots,C_{i_k}\right\}$ be the $k$ nearest neighbor centroids of $C_i$, and their coordinates are substracted by $C_i$ to get the relative coordinates of each neighbor point. 
Consequently, we obtain the Euclidean space features of $C_i$ as: 
\begin{equation}
        \small
\begin{aligned}
        F_{eu_i} = \bigodot \left(f_j - f_i,f_j\right), j \in \left\{i_1,\dots,i_k \right\}
\end{aligned}
      \end{equation}
where $\bigodot$ represents the concatenation of the relative coordinates and original coordinates.
Subsequently, we concatenate the Euclidean space features $F_{eu_i}$ with the eigenvalue space features $F_{eig_i}$ to boost the saliency of centroid $C_i$:
      \begin{equation}
              \small
      \begin{aligned}
              F_i = \left(F_{eu}^i,F_{eig}^i\right), i \in \left\{1,\dots,t \right\}
      \end{aligned}
            \end{equation}
      
\subsubsection{Feature graph construction} 
The multi-space KNN clustering outputs the features of each center point together with neighbor points. 
We introduce two GNNs for both Euclidean-based features and eigenvalue-based features in parallel to learn the topological relation of the scene.  
We regard the input features as graph nodes: $D = \left\{D_1,\dots,D_t\right\}$. 
For all the nodes, we implement a shared multilayer perception Layer (MLP) and use MaxPool in neighbor domains to get the embedded features:
      \begin{equation}
              \small
      \begin{aligned}
              {D_i}' = MaxPool\left(MLP\left( D_i\right)\right), i \in \left\{1,\dots,t \right\}
      \end{aligned}
            \end{equation}
The multi-layer feature graph is constructed to get a gloabl feature map $M = \left\{M_1,\dots,M_t\right\}$. 
      
Finally, the NetVLAD layer is used to learn K-cluster centers of D-dimension descriptors, namely $\left\{C_1,\dots,C_K|C \in R^D\right\}$, and the output is a $D\times K$-dimension vector that representing the relative distances to the K cluster centers. 
A full connected layer followed by an L2 normalization layer finally compresses the vector to a 256-dimension global descriptor. 
The Lazy Quadruplet loss is employed to minimize the descriptor distance between the query and positive pairs, at the same time, maximize the distance between the query and the hardest negative pairs: 
      \begin{equation}
              \small
              \begin{aligned}
                      \mathcal{L}_Q = \max_j \left[\left(\alpha + |f_{query} - f_{pos}| - |f_{query} - f_{neg_j}|\right)_+\right] \\
                      + \max_k \left[\left(\beta + |f_{query} - f_{pos}| - |f_{neg^*} - f_{neg_k}|\right)_+\right]
              \end{aligned}
      \end{equation}
where $f$ represents the descriptor vector and $|.|$ calculates the Euclidean distance between the two elements.
$\alpha$ and $\beta$ are two margin values and $neg^*$ is another negative descriptor chosen from different queries.
       
      \begin{figure*}[htbp]
        \setlength{\abovecaptionskip}{0.cm}
        \setlength{\belowcaptionskip}{-0.cm}
                \centering
                \subfigure[00]{
                \label{Fig.sub.1}
                \includegraphics[width=0.25\textwidth]{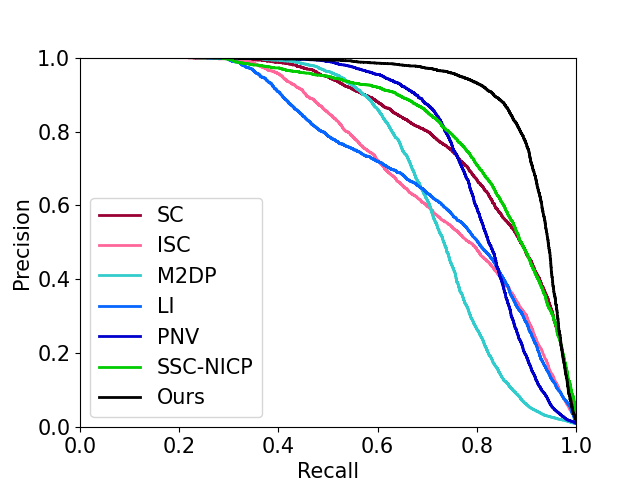}}
                \subfigure[02]{
                \label{Fig.sub.2}
                \includegraphics[width=0.25\textwidth]{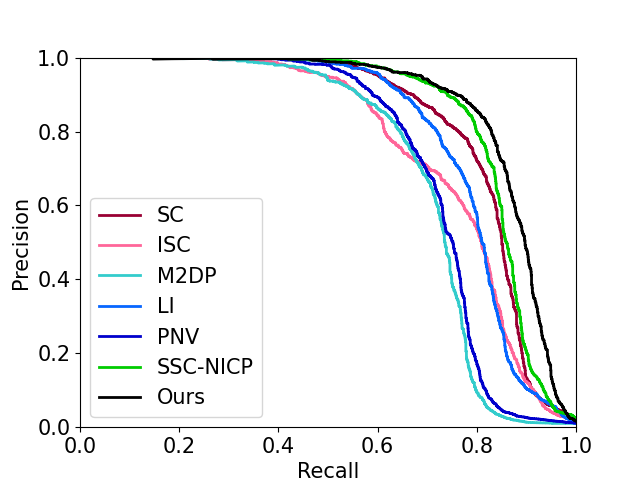}}
                \subfigure[05]{
                \label{Fig.sub.3}
                \includegraphics[width=0.25\textwidth]{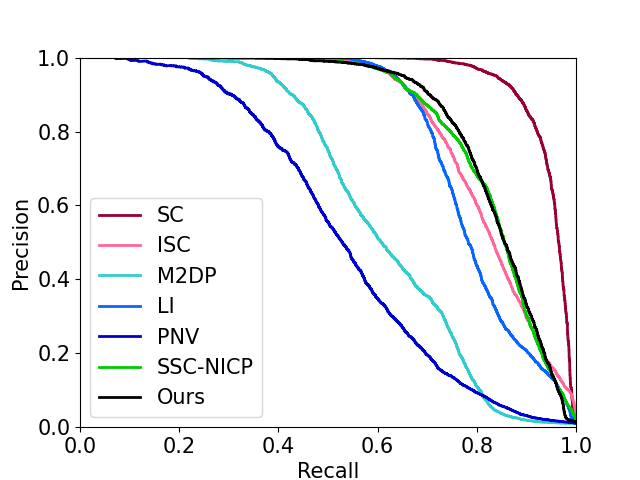}}
                \subfigure[06]{
                \label{Fig.sub.4}
                \includegraphics[width=0.25\textwidth]{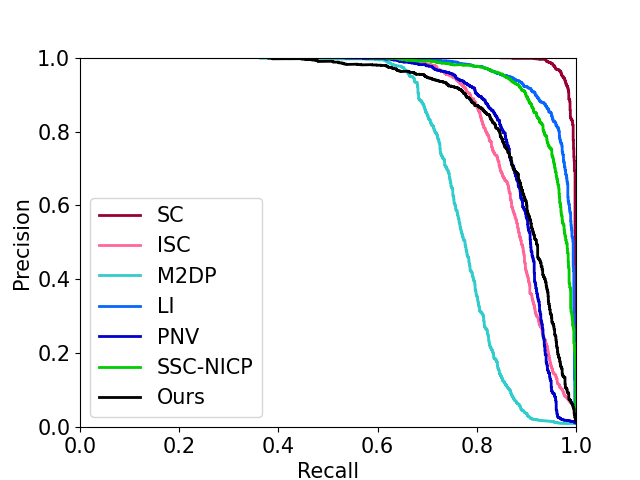}}
                \subfigure[07]{
                \label{Fig.sub.5}
                \includegraphics[width=0.25\textwidth]{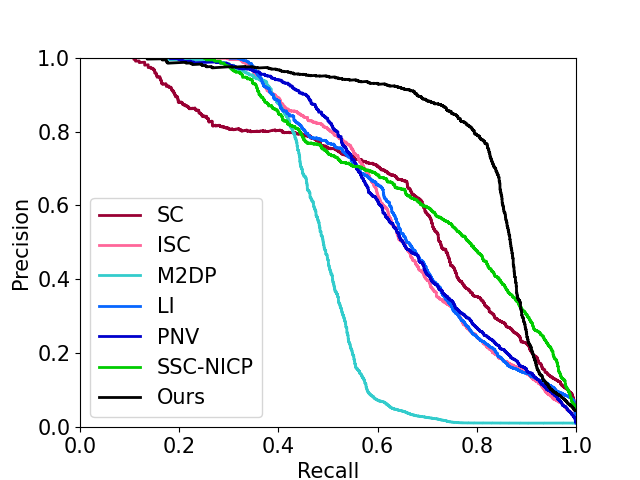}}
                \subfigure[08]{
                \label{Fig.sub.6}
                \includegraphics[width=0.25\textwidth]{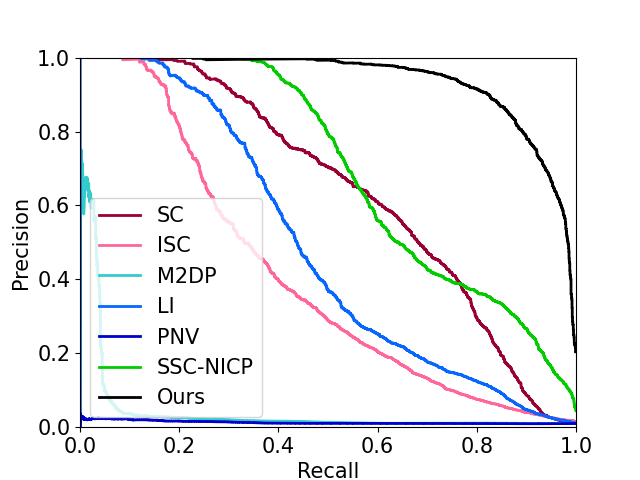}}
                \caption{Precision-Recall curves on KITTI dataset}
                \centering
                \label{PR}
                \end{figure*}
        
        \begin{table*}[tbp]
                \setlength{\abovecaptionskip}{-0pt}
                \small
                \caption{ F1 max scores and Extended Precision on KITTI dataset}
                \begin{center}
                        \begin{tabular}{cccccccc}
                                \toprule
                                \specialrule{0em}{1.2pt}{1.2pt}
                                Methods & 00 & 02 & 05 & 06 & 07 & 08 & Mean \\
                                \specialrule{0em}{1.2pt}{1.2pt}
                                \midrule
                                \specialrule{0em}{1.2pt}{1.2pt}
                                SC\cite{kim2018scan} & 0.750/0.609 & 0.782/0.632 & \textbf{0.895/0.797} & \textbf{0.968/0.924} & 0.662/0.554 & 0.607/0.569 & 0.777/0.681 \\
                                \specialrule{0em}{1.2pt}{1.2pt}
                                ISC\cite{wang2020intensity} & 0.657/0.627 & 0.705/0.613 & 0.771/0.727 & 0.842/0.816 & 0.636/0.638 & 0.408/0.543 & 0.670/0.661 \\
                                \specialrule{0em}{1.2pt}{1.2pt}
                                M2DP\cite{he2016m2dp} & 0.708/0.616 & 0.717/0.603 & 0.602/0.611 & 0.787/0.681 & 0.560/0.586 & 0.073/0.500 & 0.575/0.600 \\
                                \specialrule{0em}{1.2pt}{1.2pt}
                                LI\cite{wang2020lidar} & 0.668/0.626 & 0.762/0.666 & 0.768/0.747 & 0.913/0.791 & 0.629/0.651 & 0.478/0.562 & 0.703/0.674 \\
                                \specialrule{0em}{1.2pt}{1.2pt}              
                                PNV\cite{uy2018pointnetvlad} & 0.779/0.641 & 0.727/0.691 & 0.541/0.536 & 0.852/0.767 & 0.631/0.591 & 0.037/0.500 & 0.595/0.621 \\
                                \specialrule{0em}{1.2pt}{1.2pt}
                                Locus-SP & \textbf{0.889/0.752} & * & 0.832/0.733 & 0.880/0.793 & 0.701/0.637 & 0.734/0.666 & 0.807/0.716 \\
                                \specialrule{0em}{1.2pt}{1.2pt}              
                                SSC-NICP & 0.772/0.620 & 0.818/0.693 & 0.779/0.718 & 0.902/0.792 & 0.649/0.625 & 0.618/0.661 & 0.757/0.685 \\
                                \specialrule{0em}{1.2pt}{1.2pt}
                                Ours & 0.880/0.696 & \textbf{0.841/0.675} & 0.781/0.628 & 0.860/0.731 & \textbf{0.819/0.581} & \textbf{0.852/0.648} & \textbf{0.839/0.660} \\
                                \specialrule{0em}{1.2pt}{1.2pt}
                                \hline
                                \specialrule{0em}{1.2pt}{1.2pt} 
                                \hline
                                \specialrule{0em}{1.2pt}{1.2pt}               
                                ON\cite{chen2021overlapnet} & 0.869/0.555 & 0.827/0.639 & 0.924/0.796 & 0.930/0.744 & 0.818/0.586 & 0.374/0.500 & 0.790/0.637 \\
                                \specialrule{0em}{1.2pt}{1.2pt}               
                                SGPR\cite{kong2020semantic} & 0.820/0.500 & 0.751/0.500 & 0.751/0.531 & 0.655/0.500 & 0.868/0.721 & 0.750/0.520 & 0.766/0.545 \\
                                \specialrule{0em}{1.2pt}{1.2pt}               
                                Locus\cite{vidanapathirana2020locus} & 0.957/0.828 & * & 0.968/0.919 & 0.950/0.913 & 0.922/0.825 & 0.900/0.742 & 0.939/0.845 \\
                                \specialrule{0em}{1.2pt}{1.2pt}              
                                SSC\cite{li2021ssc} & 0.951/0.849 & 0.891/0.748 & 0.951/0.903 & 0.985/0.969 & 0.875/0.805 & 0.940/0.932 & 0.932/0.868 \\
                                \specialrule{0em}{1.2pt}{1.2pt}
                                Ours & 0.880/0.696 & 0.841/0.675 & 0.781/0.628 & 0.860/0.731 & 0.819/0.581 & 0.852/0.648 & 0.839/0.660 \\
                                \specialrule{0em}{1.2pt}{1.2pt}              
                                
                                \bottomrule
                        \end{tabular}
                        \begin{tablenotes}
                                \footnotesize
                                The methods in the upper half of the table can be applied for large-scale place retrieval while those in the bottom half are not suitable.\\
                                $^*$ In Sequence 02, some point cloud frames are failed to generate locus descriptors because spatial and temporal correspondences are unavailable.
                                
                        \end{tablenotes}
                \end{center}
                \label{F1_EP}
        \end{table*}
        \begin{table*}[tbp]
                \setlength{\abovecaptionskip}{-0pt}
                \small
                \caption{ Robustness evaluation of F1 max scores on KITTI dataset}
                \begin{center}
                        \begin{tabular}{ccccccccccccccc}
                                \toprule
                                \specialrule{0em}{1.2pt}{1.2pt}
                                \multirow{2}{*}{\makecell[c]{Methods}} & \multicolumn{7}{c}{Rotation} & \multicolumn{7}{c}{Occlusion} \\
                                \cmidrule(lr){2-8} \cmidrule(lr){9-15} & 00 & 02 & 05 & 06 & 07 & 08 & Mean & 00 & 02 & 05 & 06 & 07 & 08 & Mean \\
                                \specialrule{0em}{1.2pt}{1.2pt}
                                \midrule
                                \specialrule{0em}{1.2pt}{1.2pt}
                                SC\cite{kim2018scan} & 0.719 & 0.734 & \textbf{0.844} & 0.898 & 0.606 & 0.546 & 0.725 & 0.724 & 0.751 & \textbf{0.845} & \textbf{0.904} & 0.616 & 0.552 & 0.732 \\
                                \specialrule{0em}{1.2pt}{1.2pt}
                                ISC\cite{wang2020intensity} & 0.659 & 0.701 & 0.769 & 0.840 & 0.629 & 0.403 & 0.667 & 0.620 & 0.686 & 0.711 & 0.812 & 0.589 & 0.387 & 0.634 \\
                                \specialrule{0em}{1.2pt}{1.2pt}
                                M2DP\cite{he2016m2dp} & 0.276 & 0.282 & 0.341 & 0.316 & 0.204 & 0.201 & 0.27 & 0.199 & 0.138 & 0.283 & 0.140 & 0.113 & 0.046 & 0.153 \\
                                \specialrule{0em}{1.2pt}{1.2pt}
                                LI\cite{wang2020lidar} & 0.667 & 0.764 & 0.772 & \textbf{0.912} & 0.633 & 0.479 & 0.705 & 0.627 & 0.710 & 0.679 & 0.859 & 0.585 & 0.383 & 0.641 \\
                                \specialrule{0em}{1.2pt}{1.2pt}              
                                PNV\cite{uy2018pointnetvlad} & 0.083 & 0.090 & 0.490 & 0.094 & 0.064 & 0.086 & 0.151 & 0.547 & 0.570 & 0.295 & 0.589 & 0.444 & 0.031 & 0.413 \\
                                \specialrule{0em}{1.2pt}{1.2pt}
                                Ours & \textbf{0.852} & \textbf{0.793} & 0.737 & 0.847 & \textbf{0.775} & \textbf{0.842} & \textbf{0.808} & \textbf{0.820} & \textbf{0.773} & 0.704 & 0.834 & \textbf{0.751} & \textbf{0.800} & \textbf{0.780} \\
                                \specialrule{0em}{1.2pt}{1.2pt}           
                                \bottomrule
                        \end{tabular}
                \end{center}
                \label{robust}
        \end{table*}

\section{EXPERIMENTS}

\subsection{Experiment Setup}
For a fair comparison, we follow the experiment settings of SSC \cite{li2021ssc} on the KITTI odometry dataset, which includes 11 sequences (00-10) with ground truth poses. 
Sequence (00, 02, 05, 06, 07, 08) containing real loop closure are chosed for evaluation, and the rest sequences are used for training \cite{kong2020semantic}. 
If the distance between two point clouds is less than 3 meters or more than 20 meters, we treat the pair as positive pair or negative pair relatively.
In the evaluation, pairs whose timestamp offset is less than 30 seconds will be excluded even if they are close.

In our experiments, we set $N_s = 60$, $N_r = 20$, $L_{max} = 60 m$, so that we have 1200 segments per frame.
All experiments are implemented on the same system with an Intel Xeon E5-2680 v4 @ 2.40GHz CPU and an NVIDIA GeForce RTX 2080 Ti GPU.

\subsection{Place Recognition Performance}
We compared our method with the state-of-the-art methods including Scan Context (SC) \cite{kim2018scan}, Intensity Scan Context (ISC) \cite{wang2020intensity}, 
M2DP \cite{he2016m2dp}, LiDAR Iris (LI) \cite{wang2020lidar}, PointNetVLAD \cite{uy2018pointnetvlad}(PNV), OverlapNet (ON) \cite{chen2021overlapnet}, SGPR \cite{kong2020semantic}, 
Locus \cite{vidanapathirana2020locus}, and SSC \cite{li2021ssc}. 
Among them, OverlapNet \cite{chen2021overlapnet} and SGPR \cite{kong2020semantic} need to recognition places pair by pair, makes them unfeasible for large-scale retrieval task when no prior pose information is provided. 
The original Locus and SSC are also time-consuming since they need a pre-match step between the query and the candidates. 
We replaced the spatiotemporal feature of Locus with spatial feature due to the inefficiency caused by temporal feature pooling. 
The modified version is named Locus-SP. 
For SSC, we kept the semantic information but dropped the fast ICP, namely SSC-NICP. 
We adopt the $100N_p$ pair list presented in SSC \cite{li2021ssc} and all methods follow the same settings so that they can be evaluated fairly. 
The maximum F1 score ($F1_{max}$) and Extended Precision \cite{ferrarini2020exploring} (EP) are taken to evaluate the performance. 
The $F1_{max}$ and EP are defined as:
\begin{equation}
        \small
\begin{aligned}
        F1_{max} = \max_\tau  2\times \frac{P_\tau\times R_\tau}{P_\tau + R_\tau} \\
        EP = \frac{P_{R0} + R_{P100}}{2}
\end{aligned}
      \end{equation}
where $P_\tau$ and $R_\tau$ is the max Precision and Recall. $P_{R0}$ is the precision with minimum recall and $R_{P100}$ is the recall with 100\% precision. 
Note that if $P_{R0}$ is less than 1, then $R_{P100}$ will be 0.

The evaluation results are shown in \Reffig{PR} and \Reftab{F1_EP}. 
Both Locus and Locus-SP fail to generate Locus descriptors for some frames in sequence 02 so they can't get the results. 
In sequence 08, where the looped path is in a reverse direction, all methods except ours have accuracy drops, which proves that our method is robust to rotation variants.
For SSC-NICP, even we keep the semantic information for matching, the results are still not good. 

The methods in the bottom half of \Reftab{F1_EP} use a retrieval strategy which limits their real-time applications. 
Our method is competitive even though it is slightly behind other methods in accuracy. 
\begin{table}[htbp]
        \setlength{\abovecaptionskip}{-5pt}
        \small
        \caption{Time retrieval efficiency in large scale setting}
        \begin{center}
                \setlength{\tabcolsep}{0.3mm}{
                \begin{tabular}{cccccc}
                        \toprule
                        \specialrule{0em}{1.2pt}{1.2pt}
                        Methods & Size & KDTree/s & One Frame/ms & Total/s & Avg/s \\
                        \specialrule{0em}{1.2pt}{1.2pt}
                        \midrule
                        \specialrule{0em}{1.2pt}{1.2pt}
                        SSC\cite{kim2018scan} & $50\times360$ & 0 & 40000 & 402.6 & 40.3 \\
                        \specialrule{0em}{1.2pt}{1.2pt}                        
                        SGPR\cite{kong2020semantic} & $\left(100,32\right)$ & 0 & 1300 & 13.8 & 1.4 \\
                        \specialrule{0em}{1.2pt}{1.2pt}               
                        Locus\cite{vidanapathirana2020locus} & 4096 & 12.1 & 0.9 & 12.11 & 1.21 \\
                        \specialrule{0em}{1.2pt}{1.2pt}              
                        SC \cite{kim2018scan} & $20\times60$ & 3.7 & 0.2 & 3.702 & 0.37 \\
                        \specialrule{0em}{1.2pt}{1.2pt}   
                        M2DP \cite{he2016m2dp} & 192 & 0.3 & 0.1 & 0.301 & 0.03 \\
                        \specialrule{0em}{1.2pt}{1.2pt}   
                        Ours & 256 & 0.5 & 0.2 & 0.502 & 0.05 \\
                        \specialrule{0em}{1.2pt}{1.2pt}              
                        \bottomrule
                \end{tabular}}
        \end{center}
        \label{time}
\end{table}

\subsection{Retrieval Efficiency}
In this part, we compared our method with SC \cite{kim2018scan}, M2DP \cite{he2016m2dp}, Locus \cite{vidanapathirana2020locus}, SGPR \cite{kong2020semantic}, and SSC \cite{li2021ssc} to evaluate the time cost for single frame and multi-frame retrieval. 
We first combined the KITTI datasets (sequence 00, 02, 05, 06, 07, 08) to construct a database that contains 18236 point cloud frames. 
Then we randomly selected 10 point clouds as our query frames and searched the top 1 candidates in the database. 
For SC \cite{kim2018scan}, M2DP \cite{he2016m2dp}, Locus \cite{vidanapathirana2020locus} and our method, a KDTree is needed to save all the candidates' descriptors. 
This will cost additional time in the first retrieval. 
For SGPR \cite{kong2020semantic} and SSC \cite{li2021ssc}, their retrieval strategy has to compare the query with all the other frames one by one. 
Both the average and total retrieval time costs are evaluated and the results are shown in Table \ref{time}. 
\subsection{Robustness}
We evaluated the robustness of our method with SC \cite{kim2018scan}, ISC \cite{wang2020intensity}, M2DP \cite{he2016m2dp}, LI \cite{wang2020lidar} and PNV \cite{uy2018pointnetvlad} on challenging situations including viewpoint changes and occlusion. 
We randomly rotated the input point clouds around the z-axis, and randomly dropped the point cloud data in a range of 30 degrees to verify the robustness to occlusion. 
The same evaluation samples are selected for evaluation.
The results are shown in \Reftab{robust}. 

\section{CONCLUSION}

In this paper, we propose a general and effective place recognition method, which can be applied in real-time applications in large-scale scenes.
We segment the point cloud egocentrically to avoid the disadvantages of Euclidean segmentation. 
We introduce novel GNN-based method to embed the features of all segments with a few salient points makes the module run fast and small. 
The method is evaluated on the KITTI dataset and it shows the competitive performance and robustness against viewpoint changes and occlusion. 
In further work, we would like to take the height value into consideration when segmenting the point cloud in order to better preserve the spatial structure.
A better representation of segments is also worth exploring since the centroid coordinates are affected by rotation variants.

\addtolength{\textheight}{-12cm}   






\bibliographystyle{IEEEtran}
\bibliography{IEEEabrv,DSC}

\begin{thebibliography}{10}
\providecommand{\url}[1]{#1}
\csname url@rmstyle\endcsname
\providecommand{\newblock}{\relax}
\providecommand{\bibinfo}[2]{#2}
\providecommand\BIBentrySTDinterwordspacing{\spaceskip=0pt\relax}
\providecommand\BIBentryALTinterwordstretchfactor{4}
\providecommand\BIBentryALTinterwordspacing{\spaceskip=\fontdimen2\font plus
\BIBentryALTinterwordstretchfactor\fontdimen3\font minus
  \fontdimen4\font\relax}
\providecommand\BIBforeignlanguage[2]{{%
\expandafter\ifx\csname l@#1\endcsname\relax
\typeout{** WARNING: IEEEtran.bst: No hyphenation pattern has been}%
\typeout{** loaded for the language `#1'. Using the pattern for}%
\typeout{** the default language instead.}%
\else
\language=\csname l@#1\endcsname
\fi
#2}}

\bibitem{durrant2006simultaneous}
H.~Durrant-Whyte and T.~Bailey, ``Simultaneous localization and mapping: part
  i,'' \emph{IEEE robotics \& automation magazine}, vol.~13, no.~2, pp.
  99--110, 2006.

\bibitem{stachniss2016simultaneous}
C.~Stachniss, J.~J. Leonard, and S.~Thrun, ``Simultaneous localization and
  mapping,'' in \emph{Springer Handbook of Robotics}.\hskip 1em plus 0.5em
  minus 0.4em\relax Springer, 2016, pp. 1153--1176.

\bibitem{cummins2008fab}
M.~Cummins and P.~Newman, ``Fab-map: Probabilistic localization and mapping in
  the space of appearance,'' \emph{The International Journal of Robotics
  Research}, vol.~27, no.~6, pp. 647--665, 2008.

\bibitem{milford2012seqslam}
M.~J. Milford and G.~F. Wyeth, ``Seqslam: Visual route-based navigation for
  sunny summer days and stormy winter nights,'' in \emph{2012 IEEE
  international conference on robotics and automation}.\hskip 1em plus 0.5em
  minus 0.4em\relax IEEE, 2012, pp. 1643--1649.

\bibitem{salti2014shot}
S.~Salti, F.~Tombari, and L.~Di~Stefano, ``Shot: Unique signatures of
  histograms for surface and texture description,'' \emph{Computer Vision and
  Image Understanding}, vol. 125, pp. 251--264, 2014.

\bibitem{park2019robust}
C.~Park, S.~Kim, P.~Moghadam, J.~Guo, S.~Sridharan, and C.~Fookes, ``Robust
  photogeometric localization over time for map-centric loop closure,''
  \emph{IEEE Robotics and Automation Letters}, vol.~4, no.~2, pp. 1768--1775,
  2019.

\bibitem{bosse2013place}
M.~Bosse and R.~Zlot, ``Place recognition using keypoint voting in large 3d
  lidar datasets,'' in \emph{2013 IEEE International Conference on Robotics and
  Automation}.\hskip 1em plus 0.5em minus 0.4em\relax IEEE, 2013, pp.
  2677--2684.

\bibitem{rusu2009fast}
R.~B. Rusu, N.~Blodow, and M.~Beetz, ``Fast point feature histograms (fpfh) for
  3d registration,'' in \emph{2009 IEEE international conference on robotics
  and automation}.\hskip 1em plus 0.5em minus 0.4em\relax IEEE, 2009, pp.
  3212--3217.

\bibitem{magnusson2009appearance}
M.~Magnusson, H.~Andreasson, A.~Nuchter, and A.~J. Lilienthal,
  ``Appearance-based loop detection from 3d laser data using the normal
  distributions transform,'' in \emph{2009 IEEE International Conference on
  Robotics and Automation}.\hskip 1em plus 0.5em minus 0.4em\relax IEEE, 2009,
  pp. 23--28.

\bibitem{rusu2010fast}
R.~B. Rusu, G.~Bradski, R.~Thibaux, and J.~Hsu, ``Fast 3d recognition and pose
  using the viewpoint feature histogram,'' in \emph{2010 IEEE/RSJ International
  Conference on Intelligent Robots and Systems}.\hskip 1em plus 0.5em minus
  0.4em\relax IEEE, 2010, pp. 2155--2162.

\bibitem{he2016m2dp}
L.~He, X.~Wang, and H.~Zhang, ``M2dp: A novel 3d point cloud descriptor and its
  application in loop closure detection,'' in \emph{2016 IEEE/RSJ International
  Conference on Intelligent Robots and Systems (IROS)}.\hskip 1em plus 0.5em
  minus 0.4em\relax IEEE, 2016, pp. 231--237.

\bibitem{kim2018scan}
G.~Kim and A.~Kim, ``Scan context: Egocentric spatial descriptor for place
  recognition within 3d point cloud map,'' in \emph{2018 IEEE/RSJ International
  Conference on Intelligent Robots and Systems (IROS)}.\hskip 1em plus 0.5em
  minus 0.4em\relax IEEE, 2018, pp. 4802--4809.

\bibitem{uy2018pointnetvlad}
M.~A. Uy and G.~H. Lee, ``Pointnetvlad: Deep point cloud based retrieval for
  large-scale place recognition,'' in \emph{Proceedings of the IEEE Conference
  on Computer Vision and Pattern Recognition}, 2018, pp. 4470--4479.

\bibitem{dube2017segmatch}
R.~Dub{\'e}, D.~Dugas, E.~Stumm, J.~Nieto, R.~Siegwart, and C.~Cadena,
  ``Segmatch: Segment based place recognition in 3d point clouds,'' in
  \emph{2017 IEEE International Conference on Robotics and Automation
  (ICRA)}.\hskip 1em plus 0.5em minus 0.4em\relax IEEE, 2017, pp. 5266--5272.

\bibitem{vidanapathirana2020locus}
K.~Vidanapathirana, P.~Moghadam, B.~Harwood, M.~Zhao, S.~Sridharan, and
  C.~Fookes, ``Locus: Lidar-based place recognition using spatiotemporal
  higher-order pooling,'' \emph{arXiv preprint arXiv:2011.14497}, 2020.

\bibitem{zhu2020gosmatch}
Y.~Zhu, Y.~Ma, L.~Chen, C.~Liu, M.~Ye, and L.~Li, ``Gosmatch:
  Graph-of-semantics matching for detecting loop closures in 3d lidar data,''
  in \emph{2020 IEEE/RSJ International Conference on Intelligent Robots and
  Systems (IROS)}.\hskip 1em plus 0.5em minus 0.4em\relax IEEE, 2020, pp.
  5151--5157.

\bibitem{kong2020semantic}
X.~Kong, X.~Yang, G.~Zhai, X.~Zhao, X.~Zeng, M.~Wang, Y.~Liu, W.~Li, and
  F.~Wen, ``Semantic graph based place recognition for 3d point clouds,'' in
  \emph{2020 IEEE/RSJ International Conference on Intelligent Robots and
  Systems (IROS)}.\hskip 1em plus 0.5em minus 0.4em\relax IEEE, 2020, pp.
  8216--8223.

\bibitem{li2021ssc}
L.~Li, X.~Kong, X.~Zhao, T.~Huang, and Y.~Liu, ``Ssc: Semantic scan context for
  large-scale place recognition,'' \emph{arXiv preprint arXiv:2107.00382},
  2021.

\bibitem{mur2015orb}
R.~Mur-Artal, J.~M.~M. Montiel, and J.~D. Tardos, ``Orb-slam: a versatile and
  accurate monocular slam system,'' \emph{IEEE transactions on robotics},
  vol.~31, no.~5, pp. 1147--1163, 2015.

\bibitem{galvez2012bags}
D.~G{\'a}lvez-L{\'o}pez and J.~D. Tardos, ``Bags of binary words for fast place
  recognition in image sequences,'' \emph{IEEE Transactions on Robotics},
  vol.~28, no.~5, pp. 1188--1197, 2012.

\bibitem{wang2020lidar}
Y.~Wang, Z.~Sun, C.-Z. Xu, S.~E. Sarma, J.~Yang, and H.~Kong, ``Lidar iris for
  loop-closure detection,'' in \emph{2020 IEEE/RSJ International Conference on
  Intelligent Robots and Systems (IROS)}.\hskip 1em plus 0.5em minus
  0.4em\relax IEEE, 2020, pp. 5769--5775.

\bibitem{shan2021robust}
T.~Shan, B.~Englot, F.~Duarte, C.~Ratti, and D.~Rus, ``Robust place recognition
  using an imaging lidar,'' \emph{arXiv preprint arXiv:2103.02111}, 2021.

\bibitem{xu2021disco}
X.~Xu, H.~Yin, Z.~Chen, Y.~Li, Y.~Wang, and R.~Xiong, ``Disco: Differentiable
  scan context with orientation,'' \emph{IEEE Robotics and Automation Letters},
  vol.~6, no.~2, pp. 2791--2798, 2021.

\bibitem{steder2010narf}
B.~Steder, R.~B. Rusu, K.~Konolige, and W.~Burgard, ``Narf: 3d range image
  features for object recognition,'' in \emph{Workshop on Defining and Solving
  Realistic Perception Problems in Personal Robotics at the IEEE/RSJ Int. Conf.
  on Intelligent Robots and Systems (IROS)}, vol.~44, 2010.

\bibitem{wang2020intensity}
H.~Wang, C.~Wang, and L.~Xie, ``Intensity scan context: Coding intensity and
  geometry relations for loop closure detection,'' in \emph{2020 IEEE
  International Conference on Robotics and Automation (ICRA)}.\hskip 1em plus
  0.5em minus 0.4em\relax IEEE, 2020, pp. 2095--2101.

\bibitem{qi2017pointnet}
C.~R. Qi, H.~Su, K.~Mo, and L.~J. Guibas, ``Pointnet: Deep learning on point
  sets for 3d classification and segmentation,'' in \emph{Proceedings of the
  IEEE conference on computer vision and pattern recognition}, 2017, pp.
  652--660.

\bibitem{liu2019lpd}
Z.~Liu, S.~Zhou, C.~Suo, P.~Yin, W.~Chen, H.~Wang, H.~Li, and Y.-H. Liu,
  ``Lpd-net: 3d point cloud learning for large-scale place recognition and
  environment analysis,'' in \emph{Proceedings of the IEEE/CVF International
  Conference on Computer Vision}, 2019, pp. 2831--2840.

\bibitem{chen2021overlapnet}
X.~Chen, T.~L{\"a}be, A.~Milioto, T.~R{\"o}hling, O.~Vysotska, A.~Haag,
  J.~Behley, and C.~Stachniss, ``Overlapnet: Loop closing for lidar-based
  slam,'' \emph{arXiv preprint arXiv:2105.11344}, 2021.

\bibitem{dube2018segmap}
R.~Dub{\'e}, A.~Cramariuc, D.~Dugas, J.~Nieto, R.~Siegwart, and C.~Cadena,
  ``Segmap: 3d segment mapping using data-driven descriptors,'' \emph{arXiv
  preprint arXiv:1804.09557}, 2018.

\bibitem{xu2020geometry}
M.~Xu, Z.~Zhou, and Y.~Qiao, ``Geometry sharing network for 3d point cloud
  classification and segmentation,'' in \emph{Proceedings of the AAAI
  Conference on Artificial Intelligence}, vol.~34, no.~07, 2020, pp.
  12\,500--12\,507.

\bibitem{weinmann2014semantic}
M.~Weinmann, B.~Jutzi, and C.~Mallet, ``Semantic 3d scene interpretation: A
  framework combining optimal neighborhood size selection with relevant
  features,'' \emph{ISPRS Annals of the Photogrammetry, Remote Sensing and
  Spatial Information Sciences}, vol.~2, no.~3, p. 181, 2014.

\bibitem{arandjelovic2016netvlad}
R.~Arandjelovic, P.~Gronat, A.~Torii, T.~Pajdla, and J.~Sivic, ``Netvlad: Cnn
  architecture for weakly supervised place recognition,'' in \emph{Proceedings
  of the IEEE conference on computer vision and pattern recognition}, 2016, pp.
  5297--5307.

\bibitem{ferrarini2020exploring}
B.~Ferrarini, M.~Waheed, S.~Waheed, S.~Ehsan, M.~J. Milford, and K.~D.
  McDonald-Maier, ``Exploring performance bounds of visual place recognition
  using extended precision,'' \emph{IEEE Robotics and Automation Letters},
  vol.~5, no.~2, pp. 1688--1695, 2020.

\end{thebibliography}

\end{document}